\documentclass[letterpaper, 10 pt, journal, twoside]{IEEEtran}

\IEEEoverridecommandlockouts                              

\usepackage{cite}
\usepackage{balance}

\usepackage{graphics} 
\usepackage{epsfig} 
\usepackage{mathptmx} 
\usepackage{times} 
\usepackage{amsmath} 
\usepackage{amssymb}  
\usepackage{xcolor}
\usepackage{colortbl}
\usepackage[breaklinks,colorlinks]{hyperref}
\usepackage{caption}
\usepackage{subcaption}
\usepackage{multirow}
\let\labelindent\relax
\usepackage{enumitem}
\usepackage{booktabs}
\usepackage{physics}
\usepackage[normalem]{ulem}
\DeclareMathOperator*{\argmin}{arg\,min}
\newcommand{\ie}{{\it{i.e.}}}

\definecolor{mygreen}{rgb}{0.0, 0.5, 0.0}
\newcommand{\sfm}{{{S\emph{f}M}}}

\newcommand{\formattedparagraph}[1]{\noindent \textbf{#1}}
\newcommand{\revised}[1]{\textcolor{black}{#1}}

\begin{document}

\title{Uncertainty-Driven Dense Two-View \\ Structure from Motion}

\author{Weirong Chen, Suryansh Kumar$^\dag$,~\IEEEmembership{Member, IEEE}, Fisher Yu,~\IEEEmembership{Member, IEEE}
\thanks{Manuscript received: September, 21, 2022; Revised December, 20, 2022; Accepted January, 19, 2023. This paper was recommended for publication by Editor Cesar Cadena Lerma upon evaluation of the Associate Editor and Reviewers' comments. 
}
\thanks{All the authors are with VIS Group at ETH Z\"urich, 8092 Zürich, Switzerland (email: wrchen530@gmail.com; k.sur46@gmail.com; i@yf.io).}%
\thanks{$^\dag$Corresponding Author: Suryansh Kumar.}
}



\maketitle
\begin{abstract}
This work introduces an effective and practical solution to the dense two-view structure from motion (\sfm)~problem. One vital question addressed is how to mindfully use per-pixel optical flow correspondence between two frames for accurate pose estimation---as perfect per-pixel correspondence between two images is difficult, if not impossible, to establish. With the carefully estimated camera pose and predicted per-pixel optical flow correspondences, a dense depth of the scene is computed. Later, an iterative refinement procedure is introduced to further improve optical flow matching confidence, camera pose, and depth, exploiting their inherent dependency in rigid \sfm. The fundamental idea presented is to benefit from per-pixel uncertainty in the optical flow estimation and provide robustness to the dense \sfm~system via an online refinement. 
Concretely, we introduce our uncertainty-driven Dense Two-View \sfm~pipeline (DTV-SfM), consisting of an uncertainty-aware dense optical flow estimation approach that provides per-pixel correspondence with their confidence score of matching; a weighted dense bundle adjustment formulation that depends on optical flow uncertainty and bidirectional optical flow consistency to refine both pose and depth; a depth estimation network that considers its consistency with the estimated poses and optical flow respecting epipolar constraint. Extensive experiments show that the proposed approach achieves remarkable depth accuracy and state-of-the-art camera pose results superseding SuperPoint and SuperGlue accuracy when tested on benchmark datasets such as DeMoN, YFCC100M, and ScanNet. Code and more materials are available at \url{http://vis.xyz/pub/dtv-sfm}.
\end{abstract}


\section{Introduction}\label{sec:intro}
The dense two-view structure from motion (\sfm)~problem generally aims at recovering the relative camera motion between two frames and the per-pixel 3D position of a rigid scene. A reliable solution to this problem can be helpful in several visual automation, and robotics applications 
\cite{forster2014svo}\cite{mur2015orb}, 
since it forms the basic building block of many large-scale 3D reconstruction pipelines \cite{agarwal2011building}\cite{schonberger2016structure}. Unfortunately, as of today, the credible \sfm~systems are confined to sparse 3D reconstruction \cite{schonberger2016structure}. Yet, modern applications in mixed reality \cite{mossel2016streaming}, robot vision \cite{teed2021droid}\cite{menini2021real}, motion-capture \cite{kumar2016multi}\cite{kumar2022organic} and others wish for a reliable dense 3D acquisition and camera pose estimation from images.

One of the leading reasons for the absence of a dense two-view \sfm ~algorithm in practice, contrary to the sparse one, is that it is relatively simple to analyze a few pixel correspondences,  validate the matching accuracy, and be scalable. In other words, sparse pixel correspondence can be considered conditionally independent, making it statistically convenient to justify its reliability in camera pose and sparse 3D reconstruction \cite{fischler1981random}\cite{lowe2004distinctive}. As a result, it is easy to use for multi-view cases.
On the contrary, modeling per-pixel correspondences between two images for solving dense \sfm ~are often challenging \cite{ranjan2019competitive}. Practically, modeling per-pixel matching and assessing its consistency with camera pose and scene depth is \revised{non-trivial}. It can seriously influence the camera pose and depth estimation accuracy if not modeled aptly by the algorithm. Also, we must be careful about the pixel correspondence selection for pose estimation, as five distinct correct correspondences are theoretically sufficient for fast and accurate camera pose estimation \cite{nister2004efficient}.
Additionally, it is relatively challenging---both computationally and algorithmically---to extend dense two-view \sfm~for multi-view cases. Yet, the dense two-view case shall serve as a foundational pipeline for multi-view cases hence, important for research.

Meanwhile, recent deep-learning approaches can address some of the limitations of dense \sfm~pipelines by learning the camera motion and scene geometry priors in a supervised setting. In this regard, some recent works focus on improving the pixel correspondence quality with deep neural networks (DNNs). Their key motivation is to learn suitable feature representations and reliable matching from large-scale datasets~\cite{teed2020raft}\cite{truong2021learning}\cite{shen2020ransac}\cite{sun2021loftr}. With neural networks being the universal function approximators~\cite{hornik1989multilayer}, another trend is to overlook the geometric two-view \sfm~pipeline \cite{longuet1981computer} and directly regress the relative camera pose through the neural networks, which are usually trained jointly with another depth network in a supervised~\cite{ummenhofer2017demon}\cite{wei2020deepsfm} or self-supervised setting~\cite{zhou2017unsupervised}\cite{yin2018geonet}\cite{mahjourian2018unsupervised}. While these learning-based methods show encouraging results, most are not careful about the pixel correspondence reliability for pose estimation and how the predicted poses can affect the overall depth estimation. Even state-of-the-art deep learning-based methods mostly use well-known supervised learning pipelines and ignore the measure of correctness of the predicted pose and its impact on depth estimates~\cite{wei2020deepsfm}\cite{wang2021deep}. These matters, or even the confidence of dense correspondence prediction, have often not been considered fully in deep learning-based dense two-view \sfm~problems \cite{wang2021deep}. To mitigate the aforementioned challenges and resolve the prevailing issues in solving this problem, we propose a simple, accurate, and systematic learning-based dense two-view \sfm ~approach.

\begin{figure*}
     \centering
     \begin{subfigure}[b]{0.30\textwidth}
         \centering
         \includegraphics[width=\textwidth]{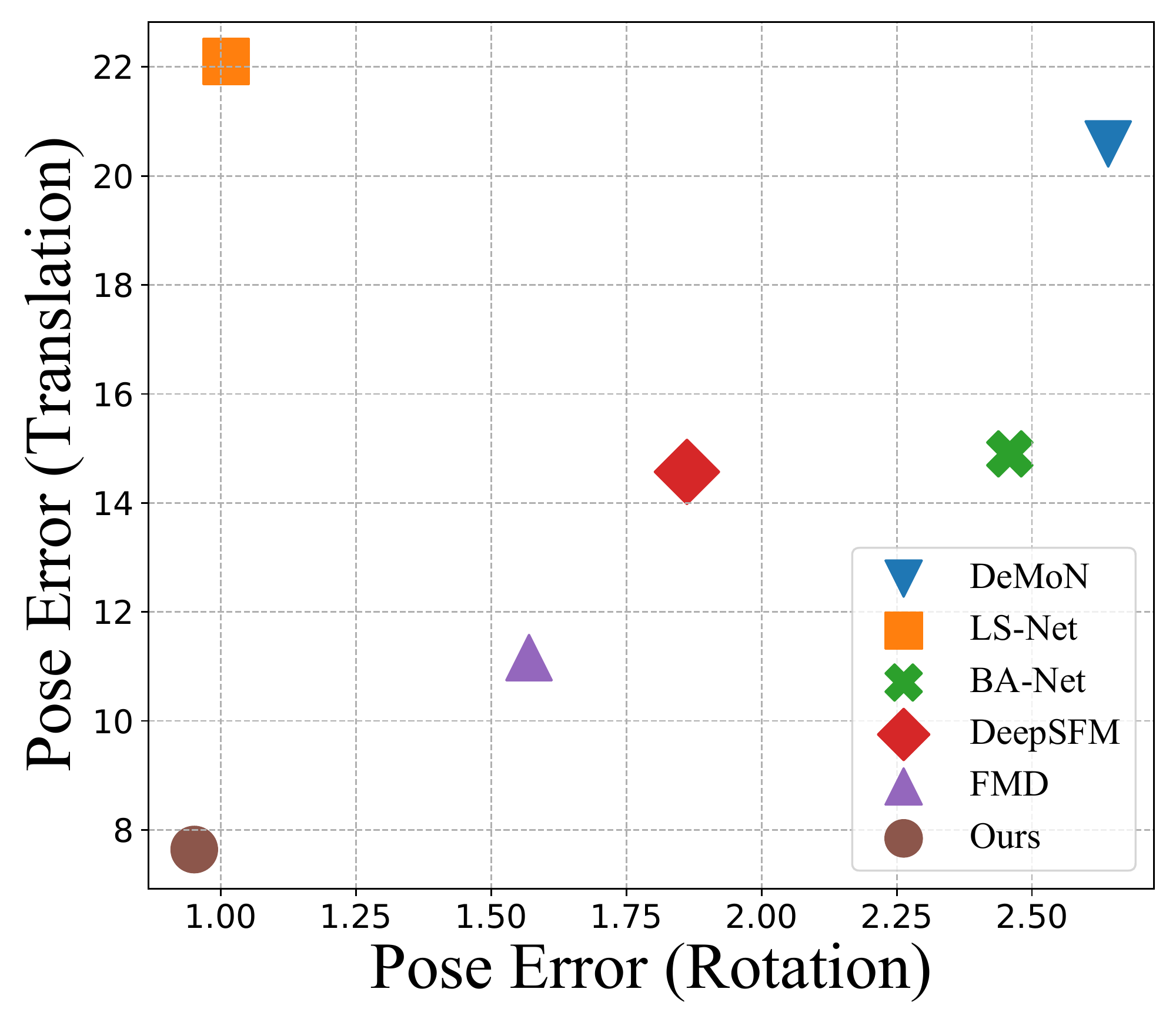}
         \caption{Camera Pose Accuracy}
         \label{fig:teaser_pose}
     \end{subfigure}
     \hfill
     \begin{subfigure}[b]{0.30\textwidth}
         \centering
         \includegraphics[width=\textwidth]{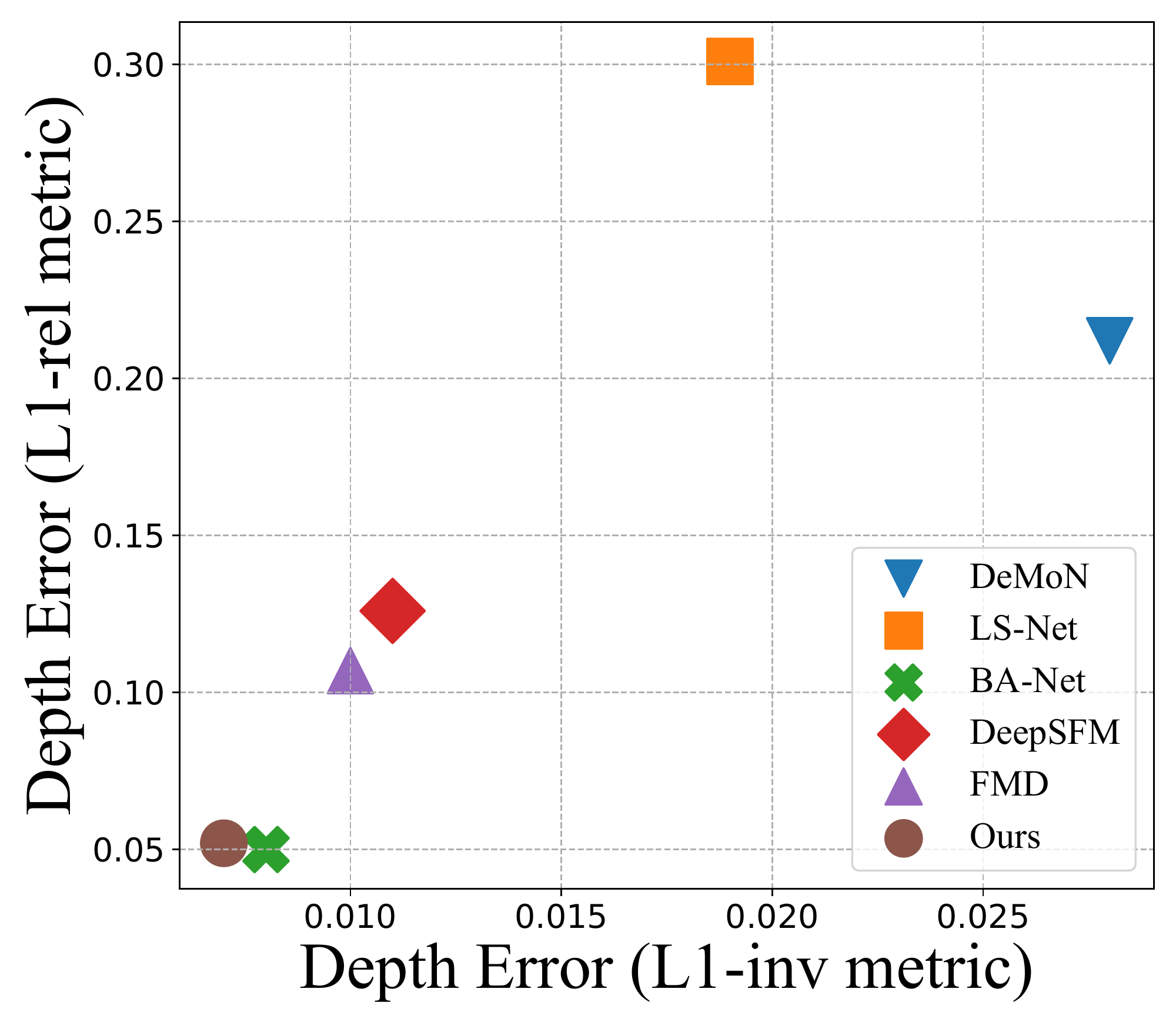}
         \caption{Depth Accuracy}
         \label{fig:teaser_depth}
     \end{subfigure}
     \hfill
     \begin{subfigure}[b]{0.30\textwidth}
         \centering
         \includegraphics[width=\textwidth]{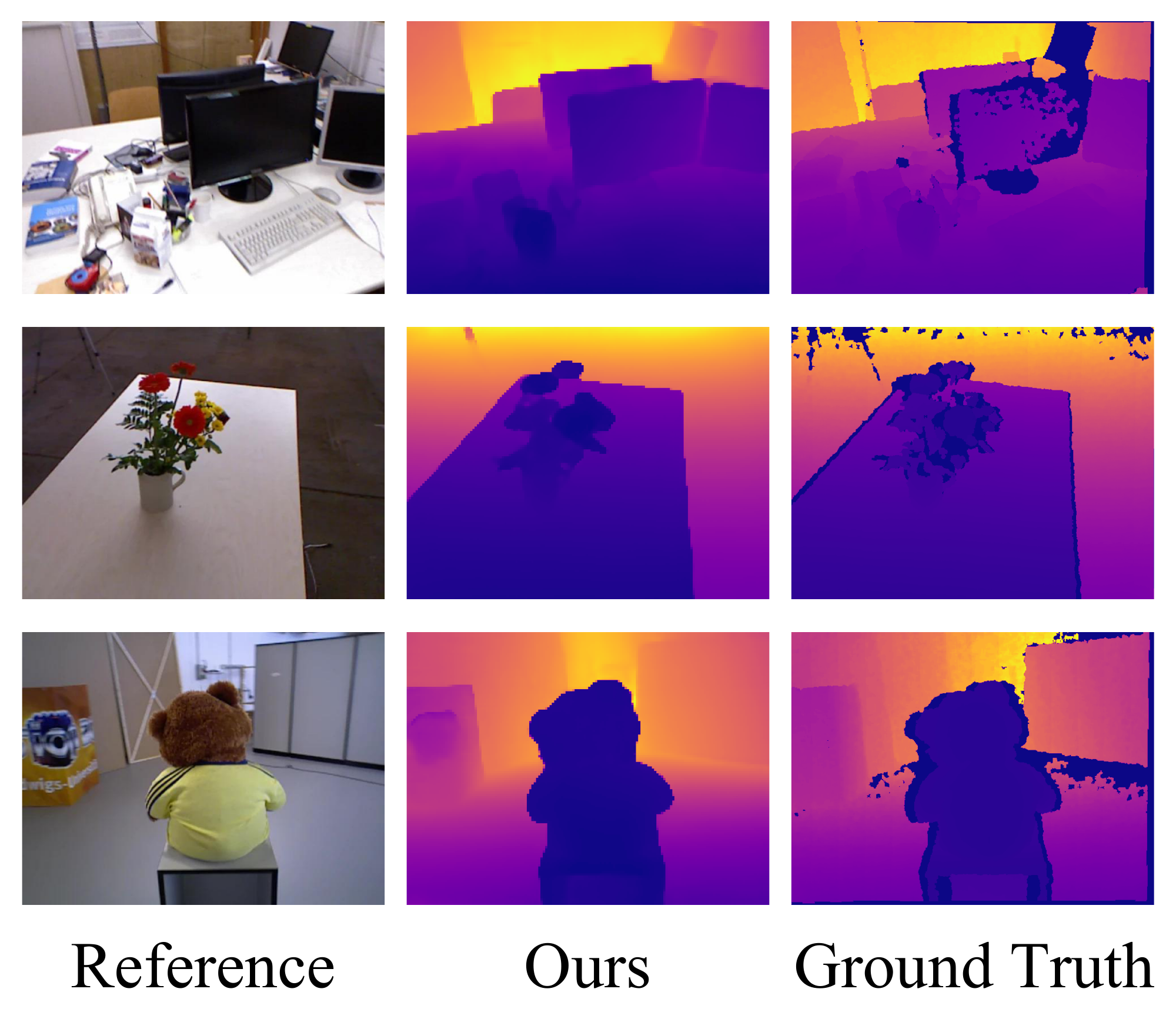}
         \caption{Qualitative Results}
         \label{fig:teaser_vis}
     \end{subfigure}
        \caption{\small Comparison with state-of-the-art learning-based dense two-view \sfm ~methods \cite{ummenhofer2017demon}\cite{wei2020deepsfm}\cite{clark2018ls}\cite{tang2018ba}\cite{wang2020flow} on RGBD dataset~\cite{sturm2012benchmark}. 
        (a)-(b) Our approach shows significant boosts in the camera pose accuracy results; at the same time, we have high-quality depth estimation results. (c) Qualitative depth estimation results compared to the respective ground truth depth map.}
        \label{fig:teaser}
\end{figure*}

The proposed approach benefits from both the classical multi-view geometry theory and learning-based confidence modeling. For camera pose estimation with known intrinsics, we introduce an uncertainty-aware pose estimation module that considers matching confidence and robust outlier filtering for camera pose recovery at test time. On one hand, with the advancement of learning-based uncertainty modeling, it is easy and convenient to model the matching uncertainty in the network design itself rather than optimize it separately.
On the other hand, conventional methods like RANSAC~\cite{fischler1981random} and other consensus set maximization methods~\cite{bazin2014globally}\cite{yang2014optimal} can perform outlier rejection and robust pose recovery, especially for the large baseline case where optical flow estimation is unreliable. 
As we demonstrate statistically later in the paper, \revised{these two methods complement each other and provide state-of-the-art camera pose estimation results when tested on a diverse benchmark dataset (see Fig.\ref{fig:teaser} (a)).}

For depth estimation, existing methods typically build a cost volume to aggregate multi-view features given the camera pose and later apply multiple 3D convolutions to obtain the depth \cite{wei2020deepsfm}\cite{wang2021deep}. On the contrary, our approach utilizes monocular and cross-view image feature cues and their consistency with the estimated poses and per-pixel matching, respecting the well-founded epipolar constraint. Furthermore, most of the existing deep-learning methods, if not all, compile their results once the camera pose and depth of the scene are recovered \cite{wang2021deep}\cite{clark2018ls}\cite{wang2020flow}. Nonetheless, to enhance the performance, in this work, we propose taking the learning-based dense \sfm ~pipeline a step further.
Notably, our work shows that we can improve camera pose and depth results by exploiting the dependency of optical flow, depth, and camera pose in rigid \sfm. This dependency loop of per-pixel correspondence confidence score, pose, and depth is encapsulated in an iterative optimization and solved until convergence. To put it simply, we compute per-pixel correspondences with the predicted pose and depth by projecting the 3D points onto the other image. We call it ``induced optical flow''. The induced optical flow provides additional evidence about the dense correspondences between frames. We update the earlier predicted optical flow and flow confidence by checking the consistency between the induced and predicted optical flow. The updated flow and flow matching confidence provides better \revised{cues} for our weighted bundle adjustment formulation---refer to Sec.\ref{ssec:WBA}, which helps refine the camera pose and depth estimates.

\noindent
\textbf{Contributions.} To summarize, our key contributions are:
\begin{itemize}[leftmargin=*]

    \item This paper proposes an accurate and reliable dense two-view \sfm~system. The proposed method utilizes the fundamentals of rigid \sfm ~with mindful statistical modeling of per-pixel optical flow matching leading to state-of-the-art camera pose estimates and favorable scene depth recovery.
    
    \item It is demonstrated that excellent camera pose can be recovered by using uncertainty-aware dense optical flow correspondence in addition to the popular statistical outlier rejection approach. Hence, we demonstrate that flow based uncertainty-driven weighted bundle adjustment (WBA) with outlier filtering and bidirectional flow consistency is a simple yet effective tool for camera pose recovery.

    \item If not all, most methods conclude their results once camera pose and depth are estimated. To this end, an iterative update scheme is introduced to improve the estimated camera pose and depth further. This later stage update is done by refining the dense optical flow and flow confidence via induced optical flow consistency. 
\end{itemize}

\section{Related Work}
Classical two-view \sfm~has been extensively studied over the past few decades. While a robust, practical, and reliable two-view \sfm~system is still limited to a sparse set of image points, a consistent effort to extend it to a dense \sfm ~system can be observed, especially with the advancement in deep neural network architectures \cite{wei2020deepsfm}\cite{teed2018deepv2d}. Thus, we survey the existing works by dividing them into classical and deep learning-based methods for better understanding.

\smallskip
\formattedparagraph{Classical two-view \sfm.} These approaches follow the well-founded projective geometry rules and formulation developed in the early 1980s-1990s \cite{longuet1981computer}. A standard pipeline for two-view \sfm~is \textit{\textbf{(a)}} for each image compute the features and descriptors; \textit{\textbf{(b)}} find the descriptor matching between the two images and filter the wrong matches using outlier rejection methods~\cite{fischler1981random}\cite{bazin2014globally}\cite{yang2014optimal}; \textit{\textbf{(c)}} estimate camera-pose and 3D structure using those matches \cite{nister2004efficient}. Such a pipeline has developed over the years and put to use in many real-world applications, mainly due to the advancement in feature descriptor \cite{lowe2004distinctive} \cite{bay2006surf}, optical flow matches \cite{teed2020raft}\cite{kumar2019superpixel}, improved optimization techniques \cite{agarwal2011building} and scalable data structures \cite{schonberger2016structure}. Since the classical two-view \sfm ~pipeline depends on well-behaved imaging conditions, its performance can degrade drastically for texture-less, specular, and non-diffuse \revised{objects} in the scene. Recent developments in deep neural networks can help resolve such a limitation with the classical pipeline, which brings us to our next discussion on two-view \sfm~methods.

\smallskip
\formattedparagraph{Deep learning for \sfm.} For better insight, we briefly discuss the key deep learning methods under two sub-categories.

\smallskip
\noindent
\textit{\textbf{(i)} Self-supervised methods.} These methods estimate the camera pose and depth without using ground truth supervision at train time. Here, the common approach is to use the view synthesis-based image consistency loss function to train the model, \ie,  one should be able to reconstruct the next image from the previous image, given the correct camera parameters,  pose, and depth map \cite{garg2016unsupervised}. Self-supervised methods generally adopt a monocular depth estimation network and a pose regression network that can be trained with the monocular image sequence alone \cite{ranjan2019competitive}\cite{zhou2017unsupervised}\cite{yin2018geonet}\cite{mahjourian2018unsupervised}. However, self-supervised methods rely heavily on imaging priors, the correctness of the predicted pose and therefore, often have difficulty handling challenging environments.

\smallskip
\noindent
\textit{\textbf{(ii)} Supervised methods.}
Contrary to self-supervised approaches, these methods rely on ground truth scene depth and camera poses at train time. By now, it is widely accepted that the key to estimating better camera pose and depth largely depends on per-pixel correspondence accuracy. Following this, several works focus on improving correspondences through learning-based methods \cite{truong2021learning}\cite{shen2020ransac}\cite{sun2021loftr}\cite{wiles2020d2d}\cite{wang2020learning}\cite{huang2022neuralmarker}. However, it is difficult for a deep-learning framework to directly adopt the classical \sfm ~idea of using a robust outlier rejection algorithm for robust camera pose estimation. Consequently, pose regression networks have been proposed in the past, which are fully differentiable by design and can be trained end-to-end. DeMoN \cite{ummenhofer2017demon} is one of the early methods that use a multi-scale encoder-decoder network to regress camera pose and depth map and iteratively refine them using dense optical flow estimates. Likewise, DeepSFM~\cite{wei2020deepsfm} transforms the camera pose and depth into two 3D volumes and performs iterative updates within the cost volume space. Some recent works also try to encode the optimization steps into the neural network as differentiable solvers. LS-Net~\cite{clark2018ls} uses a deep neural network to predict the camera pose and depth by minimizing the photometric error, whereas BA-Net~\cite{tang2018ba} uses the Levenberg-Marquardt algorithm for bundle adjustment as a differentiable layer with learnable damping factor and optimizes with the feature-metric error. More recently, DROID-SLAM~\cite{teed2021droid} proposes an end-to-end system by integrating flow, confidence, and geometric optimization via an iterative GRU update module to solve monocular dense SLAM tasks.

\section{Method}

\noindent
\textbf{Problem Statement}. Given a consecutive pair of monocular images $\mathbf{X} = (\mathbf{I}^r, \mathbf{I}^s)$ and the intrinsic camera calibration matrix $\mathbf{K} \in \mathbb{R}^{3 \times 3}$, our goal is to estimate the relative camera pose $\mathbf{T} \in SE(3)$ between those frames and a depth map $\mathbf{D}^r \in \mathbb{R}^{H \times W}$ corresponding to $\mathbf{I}^r$. Here, $H$ and $W$ symbolize the image's height and width. We denote $\mathbf{I}^r\in \mathbb{R}^{H\times W\times 3}$ and $\mathbf{I}^s \in \mathbb{R}^{H\times W\times 3}$ as the reference image and source image, respectively.

Following the widely accepted \sfm~pipeline \cite{schonberger2016structure}, we first seek an accurate per-pixel correspondence between the two frames. As alluded to above, it is a challenging task. Thus, we turn to modern deep-learning methods that have recently demonstrated outstanding results for the per-pixel correspondence problem. Nevertheless, for \sfm ~it is equally important to reason about the correctness of the estimated pixel correspondence. Therefore, we adopt a dense optical flow network with uncertainty modeling, which at test time provides a per-pixel optical flow matching $\mathbf{Y} \in \mathbb{R}^{H\times W\times 2}$ with a per-pixel confidence score matrix $\mathbf{C} \in \mathbb{R}^{H\times W}$ of the predicted matching. Our uncertainty-aware optical flow estimation network helps in excellent camera pose recovery (Sec.\ref{ssec:UAFE}).

\begin{figure*}[t]
\includegraphics[width=0.86\textwidth]{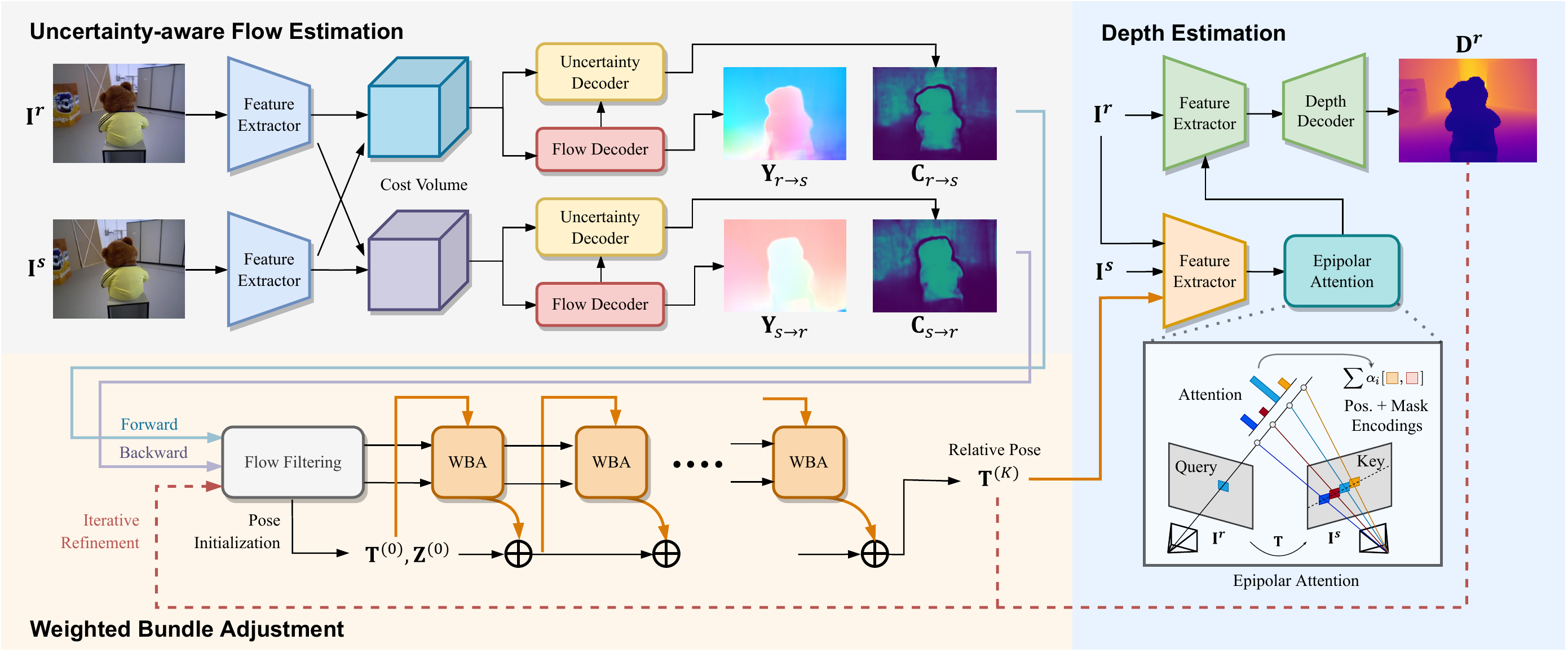}
\centering
\caption{\small An overview of our uncertainty-driven dense two-view \sfm~approach. We first predict the forward-backward dense optical flow correspondences and their confidence maps. Then, we solve the relative pose using the weighted bundle adjustment module with robust outlier rejection and bidirectional flow consistency information. Finally, we recover the dense scene depth using the predicted pose, which can update flow and flow confidence leading to further refinement of the camera pose and depth.}
\label{fig:pipeline}
\end{figure*}

For camera pose estimation, we propose weighted bundle adjustment (WBA), where weights in WBA consider flow uncertainty, robust outlier filtering, and bidirectional flow consistency (Sec.\ref{ssec:WBA}).
The recovered camera pose $\mathbf{T}$ along with $\mathbf{X}$ are passed to the depth estimation network to predict $\mathbf{D}^{r}$. Unlike other methods \cite{wang2021deep}\cite{wang2020flow}, our approach allows further improvement of $\mathbf{T}$ and $\mathbf{D}^{r}$ by iteratively updating flow $\mathbf{Y}$ and flow confidence $\mathbf{C}$ via induced optical flow due to the current estimate of $\mathbf{T}$ and $\mathbf{D}^r$ (Sec.\ref{ssec:IR}). Finally, an overall loss to train the network is presented in Sec.\ref{ssec:LF}. Next, we discuss our approach in more detail explaining the neural network design choice used in our overall pipeline (see Fig. \ref{fig:pipeline}).

\vspace{-0.2cm}
\subsection{Per-Pixel Correspondence Estimation}\label{ssec:UAFE}
Reliability of per-pixel matching from $\mathbf{I}^{r}$ to the corresponding pixel in $\mathbf{I}^{s}$ is the key to dense two-view \sfm~problem. Yet, relative camera pose estimation needs a few accurate pixel matches. Since there is no trivial way to identify the correct per-pixel correspondences directly, it motivates us to explore a neural-network-based data-driven solution to reason about the fidelity of the predicted flow, so that we can \emph{a priori} decide the suitability of the estimated correspondence. To this end, we exploit the dense correspondence method PDC-Net~\cite{truong2021learning} as our optical flow backbone. Using this backbone model, we compute dense matches relating to two frames and the pixel-wise uncertainty map representing the correspondences' reliability. As outlined in Sec.\ref{ssec:WBA}, such an idea can provide a better camera pose estimate than popular methods \cite{sarlin2020superglue}\cite{detone2018superpoint}. We introduce the associated notations with a brief recap of the PDC-Net as follows.

In the paper, we use the term forward flow and backward flow denoted as $\mathbf{Y}_{r \rightarrow s} \in \mathbb{R}^{H\times W\times 2}$ and $\mathbf{Y}_{s \rightarrow r} \in \mathbb{R}^{H\times W\times 2}$, for the optical flow estimated from $\mathbf{I}^r$ to $\mathbf{I}^s$ and vice-versa.  We represent $\mathbf{Y}_{r \rightarrow s}= F_{r\rightarrow s}(\mathbf{X};\theta)$, $\mathbf{Y}_{s \rightarrow r}= F_{s \rightarrow r}(\mathbf{X};\theta)$, where $F(\mathbf{X}; \theta)$ symbolizes the neural network parameterized by $\theta$. For notation convenience, we use the symbol $\mathbf{Y}$ for denoting optical flow in general and will explicitly denote the forward-backward flow symbol if required.

Given $\mathbf{X}$, instead of predicting the single flow vector values for each pixel, the goal is to predict the conditional probability density of the optical flow as $p(\mathbf{Y}|\mathbf{X};\theta)$. Denoting $\mathbf{y}_{ij} \in \mathbb{R}^{2 \times 1}$ and $\mathbf{\phi}_{ij} \in \mathbb{R}^{n \times 1}$ as the optical flow and predicted distribution parameter for the $(i, j)^\textrm{th}$ pixel, we write $p(\mathbf{Y}|\mathbf{X};\theta) = p(\mathbf{Y}| \Phi(\mathbf{X};\theta)) = \prod_{ij} p(\mathbf{y}_{ij}|\phi_{ij} (\mathbf{X};\theta))$. 
To estimate confidence value $c \in [0, 1]$ for a pixel $\mathbf{p}$ from the predicted distribution, the total probability of predicted optical flow within a radius $R$ of the estimated mean flow $\mathbf{\mu}$ is computed as
    $c = P(|y -\mu| < R) = \int_{\{ y\in \mathbb{R}^2: |y-\mu| < R\}} p(y|\phi)dy.$
The confidence map $\mathbf{C}=\{c_{ij}\} ~\forall ~(i, j) \in [H, W]$ can be obtained by calculating the confidence value for every pixel.

\subsection{Weighted Bundle Adjustment}\label{ssec:WBA}
With the dense correspondences, the next step is to estimate the relative camera pose $\mathbf{T}$. However, as the estimated correspondences are generally noisy and of varying quality, classical \sfm~pipelines mostly rely on off-the-shelf RANSAC algorithm~\cite{shen2020ransac} or other robust inlier maximization approaches \cite{bazin2014globally}\cite{yang2014optimal}.  
Without losing the robustness from outlier rejection, we further assess the quality of correspondence matches in the optical flow network design so that we can reason about correspondence inliers at test time.
Such a hybrid idea is not only effective for camera pose estimation but also helps refine the correspondence's measure itself. To realize such an idea, we introduce a weighted bundle adjustment module (WBA).

The proposed WBA optimizes for camera pose and depth such that the induced optical flow computed from the current estimation of camera pose and depth should be consistent with the network's predicted flow. Despite such an idea has some notion of similarity with \cite{teed2021droid} that obtains weights from convolutional layers without probabilistic modeling, we consider the uncertainty obtained from a mixture of probability models with additional robust outlier rejection, bidirectional flow consistency, careful pose initialization, and independent depth estimation, which significantly improves the accuracy of pose and depth estimates---refer to Sec.\ref{sec:experiment}.

\revised{Let $\mathbf{p}_{ij} \in \mathbb{R}^{2 \times 1}$  and $\bar{\mathbf{p}}_{ij} \in \mathbb{R}^{3 \times 1}$ denote $(i,j)^\textrm{th}$ image pixel coordinate and its homogeneous coordinate.} We compute $\mathbf{p}^s$, \ie, pixel in the source image corresponding to $\mathbf{p}^r$ in the reference image with predicted flow $\mathbf{Y}_{r \rightarrow s}$ as 
\begin{equation}
\label{eq:flow}
\mathbf{p}_\text{flow}^s = \mathbf{Y}_{r \rightarrow s}(\mathbf{p}^r) + \mathbf{p}^r.
\end{equation}

As stated, our aim of using WBA is to mindfully solve for camera pose; therefore, we propose exploiting the consistency between the induced flow and predicted flow. By induced flow, we mean the optical flow derived from the camera pose and the intermediate depth estimates. Assume $\mathbf{T} = [\mathbf{R}|\mathbf{t}]$ as the relative camera pose from $\mathbf{I}^r$ to $\mathbf{I}^s$, where $\mathbf{R}\in {SO}(3)$, and $\mathbf{t}\in \mathbb{R}^{3 \times 1}$ represents the translation vector. Let  $\mathbf{Z} = (\mathbf{Z}^r, \mathbf{Z}^s), ~\mathbf{Z}^r \in \mathbb{R}^{H\times W}, ~\mathbf{Z}^s \in \mathbb{R}^{H\times W}$ be the intermediate depth representations used in WBA corresponding to $\mathbf{I}^{r}$ and $\mathbf{I}^{s}$, respectively.
Now, we are ready to define the forward induced flow by back-projecting the pixel in $\mathbf{I}^r$ to 3D space and projecting it back to $\mathbf{I}^s$ using the $\mathbf{T}$ and $\mathbf{Z}^r$. We compute the induced-flow corresponding pixel in the source image as
\begin{equation}
\label{eq:induced_flow}
\bar{\mathbf{p}}_\textrm{ind\_flow}^s(\mathbf{T},\mathbf{Z}^r) = \mathbf{K}(\mathbf{R} \cdot \mathbf{Z}^r(\mathbf{p}^r) \cdot \mathbf{K}^{-1} \bar{\mathbf{p}}^r + \mathbf{t}).
\end{equation}
Combining Eq.\eqref{eq:flow} and Eq.\eqref{eq:induced_flow} gives us the reprojection error constraint, \ie, $\sum_{ij} ||\mathbf{p}_{ \textrm{flow},ij}^s - \mathbf{p}_{\textrm{ind\_flow}, ij}^s(\mathbf{T},\mathbf{Z}^r)||^2$ which helps refine the intermediate depth and pose. Meanwhile, we also have prior information pertaining to the quality of optical flow, \ie, the confidence map $\mathbf{C}$, which we have estimated using our uncertainty-aware flow estimation network (Sec \ref{ssec:UAFE}).
While $\mathbf{C}$ does provide self-contained evidence of the correspondence quality, we further boost the robustness by combining it with the robust outlier filtering from RANSAC. 
Therefore,
we introduce an additional binary mask $\mathbf{M}=\{m_{ij}\} ~\forall ~(i, j) \in [H, W]$ that includes high-value score $\mathbf{C}$ as well as RANSAC inliers. Accordingly, we compute the binary mask as
\begin{equation}
    m_{ij}=[\text{$c_{ij} \geq \gamma$ and $\mathbf{y}_{ij} \in A$}]
\end{equation}
where $\gamma$ is the confidence threshold, $[\cdot]$ is the Iverson bracket, and $A$ symbolizes the set of RANSAC inliers computed from $\mathbf{Y}$. Using
the defined binary mask and $c_{ij} \in \mathbf{C}$, we define the weights for our WBA as $w_{ij} = m_{ij} c_{ij}$. This leads us to our WBA formulation defined as
\begin{equation}\label{eq:fowardinducedflow}
E_{r\rightarrow s}(\mathbf{T}, \mathbf{Z}^r) = \sum_{ij} w_{r\rightarrow s, ij} ||\mathbf{p}_{\textrm{flow}, ij}^s - \mathbf{p}_{\textrm{ind\_flow},ij}^s(\mathbf{T},\mathbf{Z}^r)||^2.
\end{equation}
Eq.\eqref{eq:fowardinducedflow} is due to forward induced optical flow. Similarly, we can compute backward induced optical flow and have one more constraint $E_{s\rightarrow r}(\mathbf{T}^{-1}, \mathbf{Z}^s)$ for WBA. Combining both WBA constraints, we write the overall WBA objective as
\begin{equation}\label{eq:WBA}
\mathbf{T}, \mathbf{Z}^r, \mathbf{Z}^s = \argmin_{\mathbf{T},\mathbf{Z}^r, \mathbf{Z}^s} E_{r\rightarrow s}(\mathbf{T}, \mathbf{Z}^r) + E_{s\rightarrow r}(\mathbf{T}^{-1}, \mathbf{Z}^s).
\end{equation}
We use the Gauss-Newton method \cite{triggs1999bundle} to optimize Eq.\eqref{eq:WBA} for $K$ iterations. 
The initial pose $\mathbf{T}^{(0)}$ is obtained from the same RANSAC pose estimation step in the previous outlier filtering, and the initial depth map $\mathbf{Z}^{(0)}$ is set to all value one.
At the $k^{\text{th}}$ iteration, we compute the camera pose update $\Delta \xi^{(k)} \in \mathfrak{se}(3)~\textrm{(lie-algebra corresponding to} ~\mathbf{T})$, and the depth update $\Delta \mathbf{Z}^{(k)}$ for two depth maps. By vectorizing the camera pose and depth parameters, the updates are efficiently computed using the Schur decomposition~\cite{teed2021droid}. The camera pose and depth are then updated via retraction on $SE(3)$ and addition, respectively over the iterations as follows
\begin{equation}
\mathbf{T}^{(k+1)} = \exp(\Delta \xi^{(k)}) \circ \mathbf{T}^{(k)}, \quad \mathbf{Z}^{(k+1)} = \Delta \mathbf{Z}^{(k)} + \mathbf{Z}^{(k)}.
\end{equation}

\subsection{Depth Estimation and Iterative Refinement}\label{ssec:IR}
By now, we have detailed how we are predicting the uncertainty-aware dense optical flow and utilizing it for accurate pose estimation. In fact, those are critical to dense \sfm~nonetheless; the final goal is to recover an accurate depth map of the scene. To that end, we utilize the backbone architecture of MVS2D~\cite{yang2022mvs2d} 
with EfficientNet~\cite{tan2019efficientnet}.
Given the estimated pose $\mathbf{T}$ and $\mathbf{X}$, the proposed depth network predicts the reference image depth as $\mathbf{D}^r=G(\mathbf{T}, \mathbf{X}; \theta_D)$. However, if we switch the order of the reference and source image, \ie,  $\mathbf{X}'=(\mathbf{I}^s, \mathbf{I}^r)$, we can predict the source image depth map as $\mathbf{D}^s=G(\mathbf{T}^{-1}, \mathbf{X}'; \theta_D)$.

Additionally, we know there is an inherent dependency between camera pose, optical flow, and depth. As mentioned before (Sec. \ref{sec:intro}), most previous methods overlook exploiting this dependency to refine pose and depth. On the contrary, we propose to exploit the relation between them over iteration to improve our camera pose and depth.
The key idea is to utilize the current pose and depth estimates to update flow confidence. Given $\mathbf{T}$ and $\mathbf{D}=(\mathbf{D}^r, \mathbf{D}^s)$, we compute a new induced flow using Eq.\eqref{eq:induced_flow}, which provides additional evidence about the dense correspondences. 
The weights are penalized according to the distance between the predicted flow $\mathbf{Y}$ and the new induced flow\footnote{Alternatively, iterative reweighted least squares can also be used.}. We adopt the RBF kernel with standard deviation $\sigma$ to penalize large distances.
Taking forward flow as an example, the penalization factor can be computed as
\begin{equation}
    \Delta w_{r \rightarrow s, ij}^\textrm{iter} := \exp \left( - \frac{\Vert \mathbf{p}_{\textrm{flow},ij}^s - \mathbf{p}_{\textrm{ind\_flow},ij}^s (\mathbf{T}, \mathbf{D}^{r})\Vert ^ 2}{2\sigma^2}\right) 
\end{equation}
where the weights are updated as $ w_{r \rightarrow s, ij} := \Delta w_{r \rightarrow s, ij} \cdot w_{r \rightarrow s, ij}$.
We also update the predicted flow through a simple mixup as $\mathbf{Y}_{r \rightarrow s, ij} := \frac{1}{2} \mathbf{Y}_{r \rightarrow s,ij} + \frac{1}{2} (\mathbf{p}_{\textrm{ind\_flow},ij}^s (\mathbf{T}, \mathbf{D}^{r}) - \mathbf{p}^r_{ij})$.
We re-run the WBA with the updated flow and weight to obtain a better pose and further improve the depth.

\subsection{Loss Functions}\label{ssec:LF}
The uncertainty-aware flow estimation network is trained using the negative log-likelihood loss, while the depth estimation network is trained using standard L1 loss. 
\begin{equation}
\mathcal{L}_{flow} = -\log{p(\mathbf{Y}^{gt} | \Phi (\mathbf{X};\theta_F))}; ~\mathcal{L}_{depth} = \sum_{ij} |\mathbf{D}^{gt}_{ij} - \mathbf{D}_{ij} |.
\end{equation}

We trained the two networks separately using ground truth pose $\mathbf{T}^{gt}$ and depth $\mathbf{D}^{gt}$ at train time. The induced ground truth flow $\mathbf{Y}^{gt}$ is estimated from 
$\mathbf{T}^{gt}$  and  $\mathbf{D}^{gt}$ on-the-fly.

\section{Experiments, Results, and Ablations}
\label{sec:experiment}
\subsection{Datasets and Evaluation Metrics}
Following the previous methods \cite{wei2020deepsfm}\cite{wang2021deep}\cite{truong2021pdc}, we evaluated our proposed pipeline on four standard two-view \sfm~benchmark datasets, namely \textbf{YFCC100M} \cite{thomee2016yfcc100m}, \textbf{ScanNet} \cite{dai2017scannet},  \textbf{DeMoN} \cite{ummenhofer2017demon}, and \textbf{KITTI VO} \cite{geiger2013vision}. 

\smallskip
\noindent
\textbf{\textit{(i)} YFCC100M.} 
It is a large-scale dataset containing 100 million media objects with metadata collected from the internet. We follow the standard set-up of~\cite{zhang2019learning} and evaluate camera pose on four outdoor test scenes. Each scene contains 1000 image pairs with ground truth pose generated by \cite{heinly2015reconstructing}. 

\smallskip
\noindent
\textbf{\textit{(ii)} ScanNet.} ScanNet is a large-scale RGB-D video dataset with annotated camera poses and ground truth depth maps from 1613 scans of indoor scenes. We follow the standard set-up of~\cite{sarlin2020superglue} and evaluate our camera pose estimation method on the 1500 testing pairs.

\smallskip
\noindent
\textbf{\textit{(iii)} DeMoN.} DeMoN is a two-view \sfm~benchmark containing images from various scenes. The training set consists of images from three sources, \ie, RGBD \cite{sturm2012benchmark_rgbd}, SUN3D \cite{xiao2013sun3d}, and Scene11 \cite{ummenhofer2017demon}. RGBD is a SLAM dataset with high-quality RGB-D data with 16786 training and 160 testing pairs. Sun3D is a diverse indoor dataset with noisy camera pose and depth with 79577 training and 160 testing pairs. Scenes11 is a synthetic dataset of the simulated virtual scenes with random-positioned objects, containing 71820 training and 256 testing pairs. We use the same train split from \cite{yang2022mvs2d} for training depth.

\smallskip
\noindent \textbf{\textit{(iv)} KITTI VO.} KITTI visual odometry dataset contains image sequences with ground truth trajectories collected from real-world driving scenes. We evaluate our method on seq.09 with 1591 images and seq.10 with 1201 images.

\smallskip
\noindent
\textbf{Evaluation Metrics.} For YFCC100M and ScanNet, we evaluate the pose accuracy using the cumulative pose error curve (AUC) and mean Average Precision (mAP) of the pose error at different thresholds (5$^{\circ}$, 10$^{\circ}$, 20$^{\circ}$) \cite{truong2021pdc}. The pose error is computed as the maximum angular distance for rotation and translation vectors between ground truth and prediction.
For DeMoN, we use the standard metrics from \cite{wei2020deepsfm}.
Concretely, we assess the angular distance between rotation and translation vectors while the depth performance using scale-invariant error, L1 relative error, and L1 inverse error \cite{ummenhofer2017demon}.
For KITTI VO, we follow~\cite{zou2020learning} and report the absolute trajectory RMSE error of the entire trajectory with scale alignment. 

\subsection{Implementation Details}
We implemented our approach in Python 3.9 with PyTorch 1.11. The confidence threshold for weighted bundle adjustment is set to $\gamma=0.1$ \cite{truong2021pdc}. For RANSAC pose initialization, we adopt the OpenCV implementation with threshold 1.0 and confidence 0.99. For camera pose estimation on YFCC100M-ScanNet and KITTI VO, we use the provided models of PDC-Net+ (H)~\cite{truong2021pdc} and PDC-Net (D)~\cite{truong2021learning} pre-trained on Megadepth, respectively. For camera pose estimation on DeMoN dataset, we use the PDC-Net (D) ~\cite{truong2021learning} pre-trained on Megadepth and fine-tune on DeMoN train set for 50 epochs using Adam optimizer with $lr=5e^{-5}$. On DeMoN, for training the depth estimation network, we use the DeMoN train set and train our model for 50 epochs using Adam optimizer with $lr=8e^{-4}$, which takes 48 hours on 4 NVIDIA TITAN RTX GPUs. To maintain the stability of fine-tuned flow confidence for pose estimation, we use the additional $4\times4$ local grid filtering with confidence quantile in each local grid.

\subsection{Results}
\begin{table}[]
\caption{\small Camera pose estimation on YFCC100M~\cite{thomee2016yfcc100m}. The statistics show that our method achieves better results compared to the SOTA sparse (top) and dense (bottom) methods.}
\label{table:yfcc100m}
\scriptsize
\tabcolsep=0.18cm
\begin{tabular}{lcccccc}
\toprule
\multicolumn{1}{c}{\multirow{2}{*}{Method}} & \multicolumn{3}{c}{AUC $\uparrow$}                                                                     & \multicolumn{3}{c}{mAP $\uparrow$}                                                           \\
\multicolumn{1}{c}{}                        & \multicolumn{1}{c}{@5$^{\circ}$}    & \multicolumn{1}{c}{@10$^{\circ}$}   & \multicolumn{1}{c|}{@20$^{\circ}$}            & \multicolumn{1}{c}{@5$^{\circ}$}    & \multicolumn{1}{c}{@10$^{\circ}$}   & \multicolumn{1}{c}{@20$^{\circ}$}   \\ \midrule
SIFT~\cite{lowe2004distinctive} + ratio test                          & \multicolumn{1}{c}{24.09} & \multicolumn{1}{c}{40.71} & \multicolumn{1}{c|}{58.14}          & \multicolumn{1}{c}{45.12} & \multicolumn{1}{c}{55.81} & \multicolumn{1}{c}{67.20} \\
SIFT~\cite{lowe2004distinctive} + OANet~\cite{zhang2019learning}                                & \multicolumn{1}{c}{29.15} & \multicolumn{1}{c}{48.12} & \multicolumn{1}{c|}{65.08}          & \multicolumn{1}{c}{55.06} & \multicolumn{1}{c}{64.97} & \multicolumn{1}{c}{74.83} \\
SIFT~\cite{lowe2004distinctive} + SuperGlue~\cite{sarlin2020superglue}                            & \multicolumn{1}{c}{30.49} & \multicolumn{1}{c}{51.29} & \multicolumn{1}{c|}{69.72}          & \multicolumn{1}{c}{59.25} & \multicolumn{1}{c}{70.38} & \multicolumn{1}{c}{80.44} \\
SuperPoint~\cite{detone2018superpoint} (SP)                              & -                         & -                         & \multicolumn{1}{l|}{-}              & 30.50                     & 50.83                     & 67.85                     \\
SP~\cite{detone2018superpoint} + OANet~\cite{zhang2019learning}                                  & 26.82                     & 45.04                     & \multicolumn{1}{l|}{62.17}          & 50.94                     & 61.41                     & 71.77                     \\
SP~\cite{detone2018superpoint} + SuperGlue~\cite{sarlin2020superglue}                              & {38.72}                     & {59.13}                     & \multicolumn{1}{l|}{{75.81}} & {67.75}            & {77.41}            & {85.70}            \\ 
LoFTR-DT~\cite{sun2021loftr}                                         & \textcolor{red}{42.21}                         & \textcolor{red}{62.07}                        & \multicolumn{1}{l|}{\textcolor{red}{77.22 }}              &    \textcolor{red}{72.27}                   &     \textcolor{red}{79.99}             & \textcolor{red}{86.95}                         \\ \midrule
D2D~\cite{wiles2020d2d}                                         & -                         & -                         & \multicolumn{1}{l|}{-}              & 55.58                     & 66.79                     & -                         \\
RANSAC-Flow~\cite{shen2020ransac}                                 & -                         & -                         & \multicolumn{1}{l|}{-}              & 64.88                     & 73.31                     & 81.56                     \\
PDC-Net (D)~\cite{truong2021learning}                                  & 32.21                     & 52.61                     & \multicolumn{1}{l|}{70.13}          & 60.52                     & 70.91                     & 80.03                     \\
PDC-Net (H))~\cite{truong2021learning}                                     & 34.88                     & 55.17                     & \multicolumn{1}{l|}{71.72}          & 63.90                     & 73.00                     & 81.22                     \\
PDC-Net+ (D)~\cite{truong2021pdc}                                 & 34.76                     & 55.37                     & \multicolumn{1}{l|}{72.55}          & 63.93                     & 73.81                     & 82.74                     \\
PDC-Net+ (H)~\cite{truong2021pdc}                                 & \textcolor{blue}{37.51}            & \textcolor{blue}{58.08}            & \multicolumn{1}{l|}{\textcolor{blue}{74.50}} & \textcolor{blue}{67.35}            & \textcolor{blue}{76.56}            & \textcolor{blue}{84.56}            \\ \midrule
Ours                          & \textbf{45.48}            & \textbf{63.90}            & \multicolumn{1}{l|}{\textbf{78.05}} & \textbf{73.85}            & \textbf{80.80}            & \textbf{87.16}            \\ \bottomrule
\end{tabular}
\end{table}

\begin{table}[]
\caption{\small Camera pose estimation on ScanNet~\cite{dai2017scannet}. The statistics show that our method achieves better results compared to the SOTA sparse (top) and dense (bottom) methods. The symbol $\dagger$ represents networks that do not use ScanNet trainset.  }
\label{table:scannet}
\scriptsize
\tabcolsep=0.16cm
\begin{tabular}{lcccccc}
\toprule
\multicolumn{1}{c}{\multirow{2}{*}{Method}} &  \multicolumn{3}{c}{AUC $\uparrow$}                                                                     & \multicolumn{3}{c}{mAP $\uparrow$}                                                           \\
\multicolumn{1}{c}{}                        & \multicolumn{1}{c}{@5$^{\circ}$}    & \multicolumn{1}{c}{@10$^{\circ}$}   & \multicolumn{1}{c|}{@20$^{\circ}$}            & \multicolumn{1}{c}{@5$^{\circ}$}    & \multicolumn{1}{c}{@10$^{\circ}$}   & \multicolumn{1}{c}{@20$^{\circ}$} \\ \midrule
ORB~\cite{mur2015orb} + GMS~\cite{bian2017gms}                & 5.21           & 13.65          & \multicolumn{1}{c|}{25.36}          & -              & -              & -              \\
D2-Net~\cite{dusmanu2019d2} + NN              & 5.25           & 14.53          & \multicolumn{1}{c|}{27.96}          & -              & -              & -              \\
ContextDesc~\cite{luo2019contextdesc} + ratio test & 6.64           & 15.01          & \multicolumn{1}{c|}{25.75}          & -              & -              & -              \\
SP~\cite{detone2018superpoint} + OANet~\cite{zhang2019learning}        & 11.76          & 26.90          & \multicolumn{1}{c|}{43.85}          & -              & -              & -              \\
SP~\cite{detone2018superpoint} + SuperGlue~\cite{sarlin2020superglue}   & {16.16} & {33.81} & \multicolumn{1}{c|}{{51.84}} & -              & -              & -              \\ 
LoFTR-OT~\cite{sun2021loftr} $\dagger$       & 16.88          & 33.62          & \multicolumn{1}{c|}{50.62}          & -              & -              & -              \\
LoFTR-OT~\cite{sun2021loftr}                 & 21.51          & 40.39          & \multicolumn{1}{c|}{\textcolor{red}{57.96}}        & -              & -              & -              \\
LoFTR-DT~\cite{sun2021loftr}                 & \textcolor{red}{22.06} & \textcolor{red}{40.80} & \multicolumn{1}{c|}{57.62}          & -              & -              & -              \\ \midrule
PDC-Net $\dagger$  (D)~\cite{truong2021learning}              & 17.70          & 35.02          & \multicolumn{1}{c|}{51.75}          & 39.93          & 50.17          & 60.87          \\
PDC-Net $\dagger$  (H)~\cite{truong2021learning}              & 18.70          & 36.97          & \multicolumn{1}{c|}{53.98}          & 42.87          & 53.07          & 63.25          \\
PDC-Net+ $\dagger$  (D)~\cite{truong2021pdc}             & 19.02          & 36.90          & \multicolumn{1}{c|}{54.25}          & 42.93          & 53.13          & 63.95          \\
PDC-Net+ $\dagger$  (H)~\cite{truong2021pdc}             & \textcolor{blue}{20.25} & \textcolor{blue}{39.37} & \multicolumn{1}{c|}{\textcolor{blue}{57.13}} & \textcolor{blue}{45.66} & \textcolor{blue}{56.67} & \textcolor{blue}{67.07} \\ \midrule
Ours $\dagger$      & \textbf{24.07} & \textbf{43.58} & \multicolumn{1}{c|}{\textbf{60.35}} & \textbf{51.13} & \textbf{61.00} & \textbf{70.23} \\ 
\bottomrule
\end{tabular}
\end{table}

\formattedparagraph{\textit{(i)} Results on YFCC100M and ScanNet.}
Results for indoor and outdoor scenes are shown in  Table~\ref{table:yfcc100m} and Table~\ref{table:scannet}, respectively. Our method uses the pre-trained optical flow backbone model, which is denoted as the `PDC-Net+ (H)' \cite{truong2021pdc} in the tables. Table~\ref{table:yfcc100m}-\ref{table:scannet} statistical results indicate that our method outperforms the classic sparse keypoint-based two-view \sfm~pipeline by a significant margin, which uses SIFT descriptors~\cite{lowe2004distinctive} on both YFCC100M and ScanNet benchmark dataset. Our method further supersedes the popular state-of-the-art learning-based sparse methods, \ie, SuperPoint \cite{detone2018superpoint} + SuperGlue \cite{sarlin2020superglue} and LoFTR~\cite{sun2021loftr}, which have been commonly used by many recent 3D reconstruction and localization systems \cite{schonberger2016structure}\cite{cui2015global}. Compared to learning-based dense methods, our method outperforms the previous state-of-the-art `PDC-Net+(H)' for YFCC100M and `LoFTR-DT' for ScanNet. Noticeably, using the pre-trained optical flow backbone with our weighted bundle adjustment pose estimation shows consistent improvement on all metrics.

\begin{table*}[]
\caption{\small Depth and pose estimation results on DeMoN~\cite{ummenhofer2017demon}. 
We can observe that our method shows better results for pose estimation on all three datasets. In contrast to \cite{wei2020deepsfm} that relies on the additional pose and depth initialization for depth supervision, our depth network only uses ground-truth pose to train and achieves remarkable results.
}
\label{table:demon}
\centering
\tabcolsep=0.16cm
\renewcommand\arraystretch{1.1}
\begin{tabular}{c|ccccc|ccccc|ccccc}
\hline
\multirow{3}{*}{Method} & \multicolumn{5}{c|}{\textbf{RGBD}}                                                                      & \multicolumn{5}{c|}{\textbf{Scenes11}}                                                                  & \multicolumn{5}{c}{\textbf{Sun3D}}                                                                       \\ \cline{2-16} 
                        & \multicolumn{3}{c|}{Depth}                                            & \multicolumn{2}{c|}{Pose}       & \multicolumn{3}{c|}{Depth}                                            & \multicolumn{2}{c|}{Pose}       & \multicolumn{3}{c|}{Depth}                                            & \multicolumn{2}{c}{Pose}         \\
                        & L1-inv         & Sc-inv         & \multicolumn{1}{c|}{L1-rel}         & Rot            & Trans          & L1-inv         & Sc-inv         & \multicolumn{1}{c|}{L1-rel}         & Rot            & Trans          & L1-inv         & Sc-inv         & \multicolumn{1}{c|}{L1-rel}         & Rot            & Trans           \\ \hline
Base-SIFT               & 0.050          & 0.577          & \multicolumn{1}{c|}{0.703}          & 12.010         & 56.021         & 0.051          & 0.900          & \multicolumn{1}{c|}{1.027}          & 6.179          & 56.650         & 0.029          & 0.290          & \multicolumn{1}{c|}{0.286}          & 7.702          & 41.825          \\
Base-Matlab             & -              & -              & \multicolumn{1}{c|}{-}              & 12.813         & 49.612         & -              & -              & \multicolumn{1}{c|}{-}              & 0.917          & 14.639         & -              & -              & \multicolumn{1}{c|}{-}              & 5.920          & 32.298          \\
DeMoN \cite{ummenhofer2017demon}                   & 0.028          & 0.130          & \multicolumn{1}{c|}{0.212}          & 2.641          & 20.585         & 0.019          & 0.315          & \multicolumn{1}{c|}{0.248}          & 0.809          & 8.918          & 0.019          & 0.114          & \multicolumn{1}{c|}{0.172}          & 1.801          & 18.811          \\
LS-Net \cite{clark2018ls}                  & 0.019          & 0.090          & \multicolumn{1}{c|}{0.301}          & \cellcolor{blue!25}{1.010}          & 22.100         & 0.010          & 0.410          & \multicolumn{1}{c|}{0.210}          & 4.653          & 8.210          & 0.015          & 0.189          & \multicolumn{1}{c|}{0.650}          & 1.521          & 14.347          \\
BA-Net \cite{tang2018ba}                  & \cellcolor{blue!25}0.008          & 0.087          & \multicolumn{1}{c|}{\cellcolor{red!25}\textbf{0.050}} & 2.459          & 14.900         & 0.080          & 0.210          & \multicolumn{1}{c|}{0.130}          & 3.499          & 10.370         & 0.015          & 0.110          & \multicolumn{1}{c|}{{\cellcolor{blue!25}0.060} }          & 1.729          & 13.260          \\
FMD \cite{wang2020flow}                & 0.010          & 0.158          & \multicolumn{1}{c|}{0.107}          & 1.570          & \cellcolor{blue!25}{11.163}         & 0.015         & 0.268          & \multicolumn{1}{c|}{0.179}          & 0.615          & 5.331         & \cellcolor{red!25}{\textbf{0.009}} & 0.105          & \multicolumn{1}{c|}{0.076}          & 1.494          & 12.049          \\
DeepSFM \cite{wei2020deepsfm}               & 0.011          & \cellcolor{red!25}\textbf{0.071}
          & \multicolumn{1}{c|}{0.126}          & 1.862          & 14.570         & {0.007}          &  \cellcolor{blue!25}0.112          & \multicolumn{1}{c|}{0.064}          & 0.403          & 5.828          & 0.013          & 0.093          & \multicolumn{1}{c|}{0.072}          & 1.704          & 13.107          \\
Deep2View\cite{wang2021deep}            & -              & -              & \multicolumn{1}{c|}{-}              & -              & -              & \cellcolor{red!25}\textbf{0.005} & \cellcolor{red!25}\textbf{0.097} & \multicolumn{1}{c|}{\cellcolor{blue!25}{0.058}}          & \cellcolor{blue!25}{0.276}          & \cellcolor{blue!25}{2.041}          & \cellcolor{blue!25}{0.010}           & \cellcolor{red!25}\textbf{0.081} & \multicolumn{1}{c|}{\cellcolor{red!25}\textbf{0.057}} & \cellcolor{blue!25}{1.391}          & \cellcolor{blue!25}{10.757}          \\
Ours          & \cellcolor{red!25}\textbf{0.007} & \cellcolor{blue!25}0.086 & \multicolumn{1}{c|}{\cellcolor{blue!25}{0.052}}          & \cellcolor{red!25}\textbf{0.951} & \cellcolor{red!25}\textbf{7.640} & \cellcolor{blue!25}{0.007}          & {0.115}          & \multicolumn{1}{c|}{\cellcolor{red!25}\textbf{0.053}} & \cellcolor{red!25}{\textbf{0.107}} & \cellcolor{red!25}{\textbf{0.772}} & 0.011          & \cellcolor{blue!25}{0.092}         & \multicolumn{1}{c|}{0.062}          & \cellcolor{red!25}{\textbf{1.165}} & \cellcolor{red!25}{\textbf{9.731}} \\ 

\hline
\end{tabular}
\end{table*}

\formattedparagraph{\textit{(ii)} Results on DeMoN.}
Unlike YFCC100M and ScanNet dataset, DeMoN provides both depth and camera pose ground truth for evaluation. Table~\ref{table:demon} shows the quantitative comparison results of our method with other competing approaches. Results for both pose and depth in Table~\ref{table:demon} are shown side-by-side for better understanding. It can be inferred from the table that our method outperforms all other methods in camera pose performance accuracy.
Note that we follow \cite{wang2021deep} and rely on pre-trained models as outlined in the implementation, contrary to the DeepSFM \cite{wei2020deepsfm} and FMD \cite{wang2020flow} that train their pose network directly on the DeMoN pose set.

Unlike \cite{wei2020deepsfm} which trains depth networks using their network's predicted pose to better fit the ground-truth depth, we train the depth estimation network using ground truth pose alone. This helps assess the robustness of the depth estimation network against the predicted pose at test time. Contrary to \cite{wei2020deepsfm}, our depth-network training methodology is close to the conventional way of training a deep neural network.

\begin{table}[]
\centering
\tabcolsep=0.1cm
\renewcommand\arraystretch{1.1}
\caption{\small RMSE (m) results comparison with competing methods on KITTI VO Seq.09 and Seq.10~\cite{geiger2013vision}.  }
\label{table:kitti}
\begin{tabular}{c|c|c|c|c|c}
\hline
Method & SfMLearner~\cite{zhou2017unsupervised} & CC~\cite{ranjan2019competitive}    & DMVO~\cite{shen2019beyond} & LVMVO~\cite{zou2020learning} & Ours           \\ \hline
Seq.09 & 24.31      & 29.00 & 27.08       & 11.30 & \textbf{6.65}  \\
Seq.10 & 20.87      & 13.77 & 24.44       & 11.80 & \textbf{5.58} \\ \hline
\end{tabular}
\end{table}

\formattedparagraph{\textit{(iii)} Results on KITTI VO.} Further, we evaluate our pose accuracy on KITTI VO, as shown in Table~\ref{table:kitti}. We align all trajectories to the ground truth before computing the RMSE. Our method achieves better results than other learning-based methods without fine-tuning on the KITTI VO dataset. 

\subsection{Ablation Study}\label{ssec:ablation}
\formattedparagraph{\textit{(i)} Optical flow modeling and its effect on camera pose estimation.}
The proposed camera pose estimation method uses weighted bundle adjustment to aptly encapsulate per-pixel correspondence uncertainty, the bi-directional flow consistency, and the \revised{robust outlier rejection.}
Previous methods, for example, PDC-Net \cite{truong2021pdc} directly uses RANSAC~\cite{shen2020ransac} with the forward optical flow and \revised{fixed confidence threshold} to estimate camera pose, and hence overlooks to exploit the uncertainty information and bi-directional flow consistency mindfully. As shown in Table \ref{table:pose_ablation}, using the introduced weighted bundle adjustment that integrates both optical flow prediction confidence and outlier filtering achieves significant improvements for both single-direction and bi-direction cases.
Further, the use of bi-directional optical flow estimates outperforms single-directional flow by a clear margin, indicating the effectiveness of bi-directional optical flow consistency.

\begin{table}[]
\centering
\caption{
\small Effect of the proposed pose estimation method on ScanNet~\cite{dai2017scannet}. `F', `F-B' symbolizes forward flow and forward-backward flow, respectively.
}
\label{table:pose_ablation}
\begin{tabular}{@{}cccccc@{}}
\toprule
\multirow{2}{*}{Method} & \multirow{2}{*}{Flow} & \multirow{2}{*}{Mask} & \multicolumn{3}{c}{AUC$\uparrow$}                           \\ \cmidrule(l){4-6} 
                        &                       &                       &  @5$^{\circ}$                & @10  $^{\circ}$              & @20 $^{\circ}$             \\ \midrule
RANSAC                  & F                     & Conf.                 & 19.47          & 38.62          & 56.80          \\
Ours                    & F                     & RANSAC                 & 19.06          & 37.01          & 54.55          \\
Ours                    & F                     & Conf.                 & 20.11          & 37.67          & 55.32          \\
Ours                    & F                     & Conf.+RANSAC           & \textbf{21.82} & \textbf{40.14} & \textbf{57.59} \\ \midrule
Ours                    & F-B                   & -                  & 20.85          & 39.46          & 56.88          \\
Ours                    & F-B                   & RANSAC                 & 21.31          & 40.00          & 56.88          \\
Ours                    & F-B                   & Conf.                 & 22.62          & 40.93          & 58.02          \\
Ours                    & F-B                   & Conf.+RANSAC           & \textbf{24.07} & \textbf{43.58} & \textbf{60.35} \\ \bottomrule
\end{tabular}
\end{table}

\begin{figure}
     \centering
      \begin{subfigure}[b]{0.18\textwidth}
         \centering
         \includegraphics[width=\textwidth]{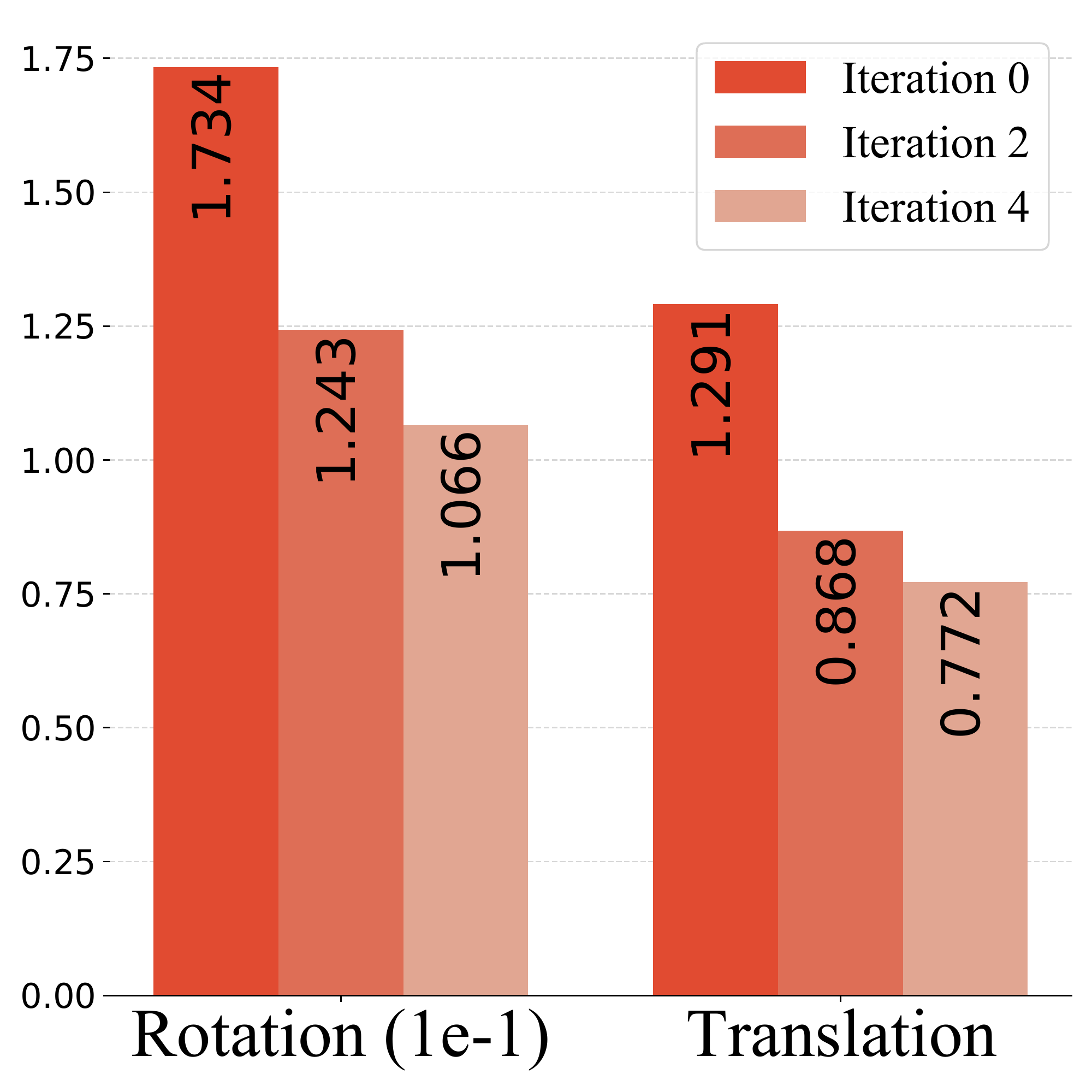}
         \caption{Pose Error}
         \label{fig:teaser_depth}
     \end{subfigure}
     \hfill
     \hfill
     \begin{subfigure}[b]{0.28\textwidth}
         \centering
         \includegraphics[width=\textwidth]{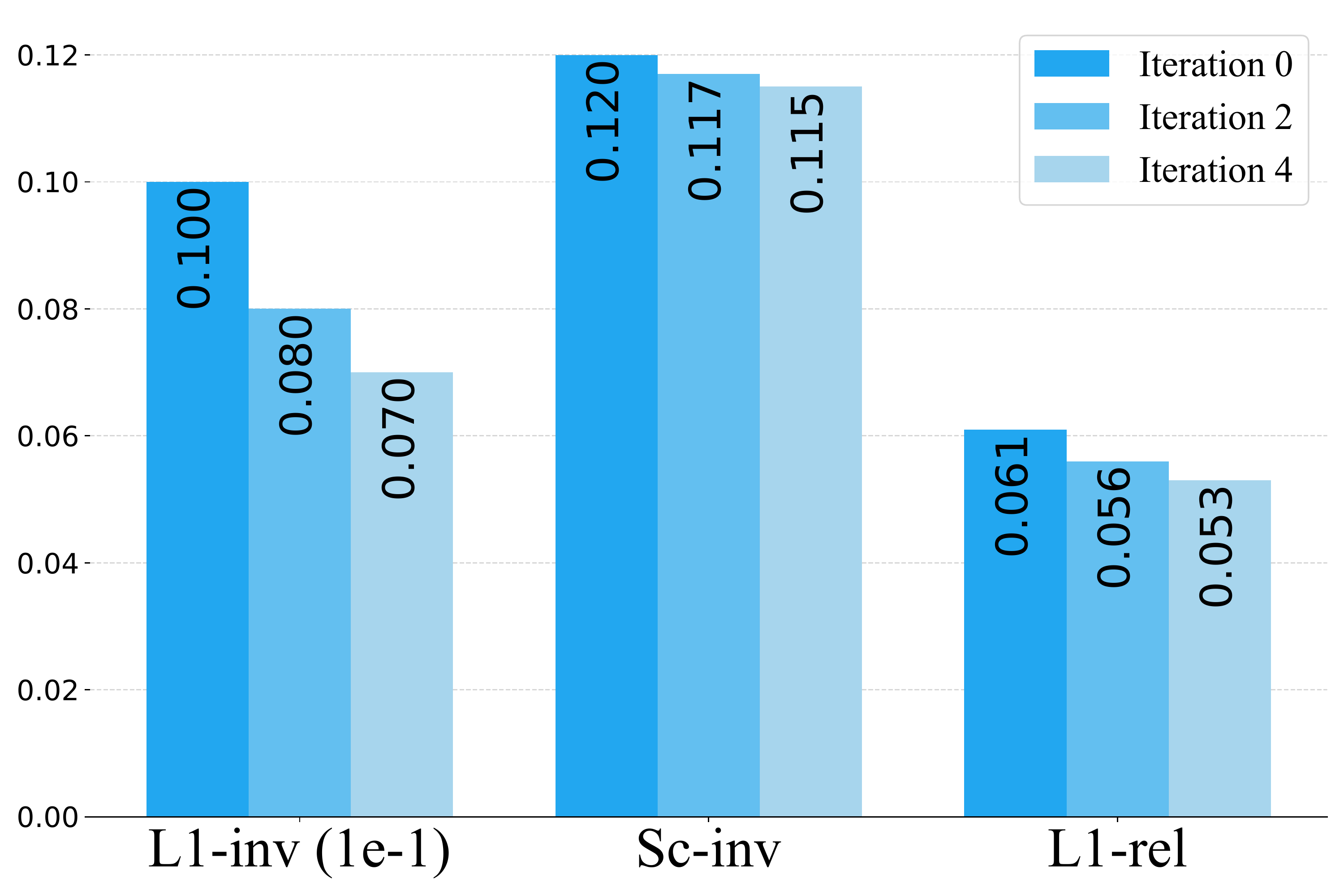}
         \caption{Depth Error}
         \label{fig:teaser_pose}
     \end{subfigure}
    \caption{
   \small 
    Iterative refinement on Scenes11~\cite{ummenhofer2017demon}. It shows that pose and depth errors can be largely reduced via iterative refinement.
    }
    \label{fig:ablation_iter}
\end{figure}

\formattedparagraph{\textit{(ii)} Use of iterative refinement.}
Here we show that we can iteratively improve camera pose and depth by exploiting the dependency loop of optical-flow correspondence, camera pose, and depth. We can induce the flow from the predicted camera pose and depth, which provides additional evidence of the flow uncertainty. In Figure~\ref{fig:ablation_iter}, we report the camera pose and depth error results for iterative refinement on Scenes11. We can observe that camera pose and depth error decrease over iterations, indicating that the joint information obtained from pose and depth estimation can provide helpful cues for updating the flow and reweighting the flow confidence. After four iterations, the rotation error drops 38\%, and the translation drops 40\%. The pose update also converges along with the depth error as the reweighting process does not obtain new information from induced flow consistency.

\begin{table}[]
\caption{\small Effect of fine-tuning on DeMoN\cite{ummenhofer2017demon}. The statistics show our approach's effectiveness for both w/ and w/o fine-tuning.}
\label{table:demon-fine-tune}
\tabcolsep=0.14cm
\renewcommand\arraystretch{1.1}
\begin{tabular}{c|cc|cc|cc}
\hline
\multirow{2}{*}{Method} & \multicolumn{2}{c|}{\textbf{RGBD}}       & \multicolumn{2}{c|}{\textbf{Scenes11}}   & \multicolumn{2}{c}{\textbf{Sun3D}}        \\ \cline{2-7} 
                        & Rot            & Trans          & Rot            & Trans           & Rot            & Trans           \\ \hline
PDC-Net (pre-trained)       & 1.311          & 10.424         & 0.721          & 5.745          & 1.382          & 13.331          \\
Ours (pre-trained)       & 0.987         & 9.754          & 0.429         & 2.684          & 1.256          & 11.079          \\ \hline
PDC-Net (fine-tuned)       & 1.108          & 8.803          & 0.321          & 2.210          & 1.292          & 12.197          \\
Ours (fine-tuned)          & \textbf{0.951} & \textbf{7.640} & \textbf{0.107} & \textbf{0.772} & \textbf{1.165} & \textbf{9.731} \\ 
\hline
\end{tabular}
\end{table}

\formattedparagraph{\textit{(iii)} Effect of fine-tuning on pose estimation.} We show the influence of fine-tuning on our pose estimation method in Table~\ref{table:demon-fine-tune}. For PDC-Net \cite{truong2021learning}, we run RANSAC on predicted optical flow to recover camera poses. Our method shows clear improvement on both pre-trained and fine-tuned models.

\formattedparagraph{\textit{(iv)} Runtime Evaluation.} 
Table~\ref{table:runtime} provides our method's runtime analysis on a single Nvidia GTX 1080 Ti GPU. Due to our modular design on the two-view SfM problem, we also have the flexibility to choose between accuracy and efficiency depending on the requirement.

\begin{table}[]
\centering
\caption{\small Runtime evaluation.}
\label{table:runtime}
\begin{tabular}{@{}ccccc@{}}
\toprule
Flow Est. & RANSAC & WBA   & Depth Est. & Total \\ \midrule
0.13s & 0.14s  & 0.21s & 0.17s & 0.65s \\ \bottomrule
\end{tabular}
\end{table}

\section{Conclusion}
From our extensive experimental analysis and rigorous ablation study, 
we conclude that our approach, which explicitly models the correctness of per-pixel correspondence matching in the neural network design itself with classical take for pose estimation, outperforms recent methods' results. Moreover, our approach for performing online refinement via weighted bundle adjustment help further improve camera pose and depth estimation accuracy. In this regard, it is observed that the proposed iterative refinement is indeed vital to exploit the dependency of optical flow, depth, and camera pose in \sfm, which is generally overlooked in the previous deep learning-based dense \sfm~approaches. That said, our approach assumes known intrinsic camera parameters, and we hope to extend the pipeline to fully uncalibrated \sfm~in our follow-up work.

\section*{Acknowledgments}
This work was supported by an ETH RobotX research grant funded through the ETH Z\"urich Foundation.

\balance
\bibliographystyle{IEEEtran}
\bibliography{IEEEabrv.bib, IEEEexample.bib}

\end{document}